# Data Poisoning and Leakage Analysis in Federated Learning

**Wenqi Wei, Tiansheng Huang, Zachary Yahn, Anoop Singhal, Margaret Loper, and Ling Liu**

## 1 Introduction

Federated learning (FL) [46] enables collaborative model training over a large corpus of decentralized data residing on a distributed population of edge clients. All clients can keep their sensitive data private and only share local model updates with the federated server.

Training data manipulation and training data privacy intrusion are two dominating threats in federated learning. Despite the default privacy by keeping client data local, recent studies [3, 21, 28, 79, 83, 86, 94, 95, 100, 101] have shown that training data leakage (usually referred to as gradient inversion or gradient leakage) intrudes client privacy because each contributing client shares its local training parameter updates (in the form of weights or gradients) with the server. By gaining access to such raw gradient information, an adversary can effectively reconstruct the private client training data by reverse engineering based on the gradients from each client at each round. In the meantime, many have exploited training data manipulation in federated learning in the form of data poisoning attacks. Data poisoning attacks

W. Wei (✉)
Georgia Institute of Technology, School of Computer Science, Atlanta, GA, USA

Fordham University, Bronx, NY, USA
e-mail: wwei23@fordham.edu

T. Huang · Z. Yahn · L. Liu
Georgia Institute of Technology, School of Computer Science, Atlanta, GA, USA

A. Singhal
National Institute of Standards and Technology, Gaithersburg, MD, USA

M. Loper
Georgia Tech Research Institute, Atlanta, GA, USA

can have two different attack objectives [8, 17, 25, 61, 71, 99] on the trained global model: (i) denial-of-service (DoS) attack such that federated training fails to converge to a point with reasonable accuracy and (ii) targeted attack such that erroneous decisions will be made only to those manipulated inputs while keeping high test accuracy on the objects of the rest.

Existing research against training data privacy intrusion relies on model perturbation by adding randomized noise to sanitize the raw gradients before sharing them with the server [83, 84, 101]. A key challenge for privacy protection by model perturbation is finding a scalable approach to determining the right amount of noise to sanitize the raw gradients while meeting the two seemingly conflicting optimization goals: The noise injected should be just enough to prevent gradient leakage inference, and yet not too much such that the negative effect on both convergence and accuracy of federated learning is minimized. Meanwhile, model perturbation is also leveraged for poisoning mitigation [7, 45, 55, 67, 76]. Similarly, it is difficult to determine the injected perturbation with the maximal mitigation on the attack effect and yet minimal negative impact on the unaltered queries.

With the increasing concerns about the data privacy and poisoning threats in federated learning, we attempt to bridge research gaps by (1) uncovering the circumstances and conditions that lead to detrimental effects from training data privacy intrusion and training data manipulation and (2) identifying the enabler and limitations of privacy protection and security assurance strategies based on model perturbation in federated learning.

To achieve these objectives, we reveal the truths and pitfalls of understanding two dominating threats: data privacy intrusion and training data manipulation. *First*, we formulate the training data leakage attacks regarding the intrinsic relationship between the training examples and their gradients. We show how adversaries can reconstruct the private local training data by simply analyzing the shared parameter update from local training (e.g., local gradient or weight update vector). We then present three observations on training data privacy leakage regarding the access of the training model, the informative gradients in early training, and the effect of model perturbation with constant noise. We compare alternative model perturbation methods, such as gradient compression, random noise injection, and differential privacy noise, concerning the proper amount and location of perturbation against training data privacy leakage. *Second*, we formulate training data manipulation attacks with targeted attack goals, which aim to cause the trained global model in federated learning only to misclassify the input from a specific victim class or with a specific pattern (trigger) into some designated malicious behavior. Then we demonstrate three observations on model access in poisoning attacks, poisoning effectiveness in terms of attack entry point, and the corresponding flaw of model perturbation with constant noise injection against training data manipulation. We analyze alternative defense approaches against training data manipulation for their mitigation effect and limitations. For both training data privacy intrusion and training data manipulation, we demonstrated the feasibility of best balancing privacy protection, poisoning resilience, and model performance with dynamic model perturbation, using dynamic differential privacy noise as the example. *At last*, we study additional risk factors of federated learning including, data skewness and misinformation. These threats exist in all learning-based systems, and their occurrence in federated learning also poses security challenges to its usability. Our analytical study with strong empirical evidence provides transformative enlightenment on effective privacy protection and security assurance strategies in federated learning, while in compliance with those trustworthy AI guidelines, such as the NIST's AI Risk Management Framework [68].

## 2 Federated Learning Preliminary

In federated learning, the machine learning task is decoupled from the centralized server to a set of $N$ client nodes. Given the unstable client availability, for each round of federated learning, only a small subset of $K_t$ clients out of all $N$ participants will be chosen to participate in the joint learning.

**Local Training at a Client** Upon notification of being selected at round $t$, a client will download the global state $w(t)$ from the server, perform a local training computation on its local dataset and the global state, i.e., $w_k(t + 1) = w_k(t) - \eta \nabla w_k(t)$, where $w_k(t)$ is the local model parameter update at round $t$ and $\nabla w$ is the gradient of the trainable network parameters. Before sharing, clients can decide its training batch size $B_t$ and the number of local iterations.

**Update Aggregation at Federated Learning Server** Upon receiving the local updates from all $K_t$ clients, the server incorporates them and updates the global state and initiates the next round of federated learning. Given that local updates can be in the form of either gradient or model weight updates, thus two update aggregation implementations are the most representative:

**Distributed SGD** At each round, each of the $K_t$ clients trains the local model with the local data and uploads the local gradients to the federated learning server. The server iteratively aggregates the local gradients from all $K_t$ clients into the global model and checks if the convergence condition of federated learning task is met. If not, the server starts the next iteration round [41, 42, 92, 93].

$$w(t + 1) = w(t) - \eta \sum_{k=1}^{K_t} \frac{n_k}{n} \nabla w_k(t),$$

where $\eta$ is the global learning rate and $\frac{n_k}{n}$ is the weight of client $k$. Here we adopt the same notation as in reference [46] so that $n_k$ is the number of data points at client $k$ and $n$ indicates the amount of total data from all participating clients at round $t$. Figure 1 provides a system overview of federated learning with distributed SGD.

**Federated Averaging** At each round, each of the $K_t$ clients uploads the local training parameter update to the federated learning server. The server iteratively performs a weighted average of the received weight parameters to update the global model and starts the next iteration round $t + 1$ unless it reaches the convergence [7, 46].

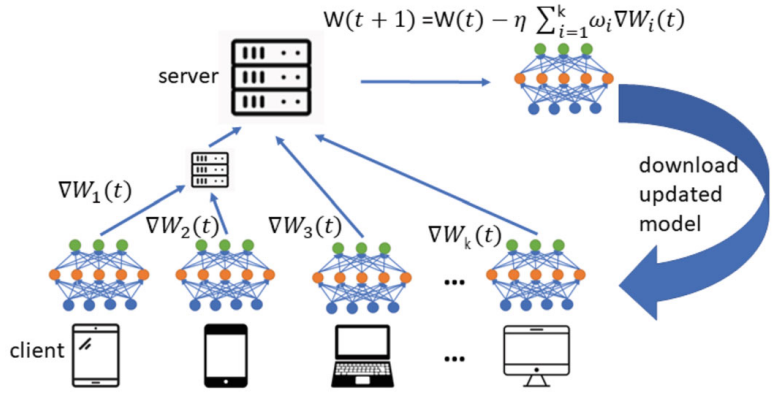

**Fig. 1** Federated learning schema

$$w(t + 1) = \sum_{k=1}^{K_t} \frac{n_k}{n} w_k(t + 1).$$

Let $\Delta w_k(t)$ denote the difference between the model parameter update before the local training and the model parameter update after the training for client $k$. Below is a variant of this method [23]:

$$w(t + 1) = w(t) + \sum_{k=1}^{K_t} \frac{n_k}{n} \Delta w_k(t).$$

## 3 Data Leakage and Privacy Protection

### 3.1 Threat Model

Training data privacy leakage is a major threat to client privacy in federated learning [21, 28, 79, 83, 94, 95, 100, 101]. The early attempt [3] brought theoretical insights by showing provable reconstruction feasibility on a single neuron or single-layer networks. Then follows the work of [21, 83, 101], [101] show the effectiveness of inverting gradients via pixel-level reconstruction to expose client training data by jointly optimizing the label and data from the dummy data to match target gradients. [83] showed that a patterned randomized attack seed can lead to a highly efficient reconstruction process in attack timing and attack effectiveness compared to the random seed. [21] demonstrated that the attack can succeed on deeper models and larger datasets with Adam optimizer. [94] propose the group consistency regularization framework that makes the gradient leakage attack on a large batch of data at ImageNet level possible.

The occurrence of training data privacy leakage in federated learning relies on several assumptions. On the data side, data at rest and data in network transit are encrypted and secure. This also implies the attackers cannot gain access to the training data prior to feeding the decrypted training data to the deep learning algorithm during local training. Therefore, the main attack surface is during data-in-use either at the client's local training or at the server's global aggregation. Training data privacy leakage usually assumes semi-curious adversary [21, 83, 101], which means the adversary may launch training data inference to reconstruct the private client training data based solely on the shared gradient updates contributed by the client.

Given the two levels of stochastic gradient descent (SGD)-based optimizations in producing a global federated model, which are server-side aggregation using FedSGD or FedAveraging and client-side training with SGD, unauthorized inference to gradient updates can happen at two possible attack surfaces. At the server side, prior to performing global aggregation of local model updates at the round $t$, the adversary may collect gradient updates from any or all of the $k_t$ participating clients and perform unauthorized reconstruction inference by model inversion, resulting in uncovering the sensitive local training data used to produce the local model update (gradients). In the rest of the chapter, we refer to such attacks as **training data leakage at server aggregation**. The adversary can also launch the training data leakage attack at a compromised client on two different gradients: (i) the accumulated per-client gradients upon completing the local model training and before encrypting it for sharing with the federated learning server or (ii) the single-step per-example gradient during each iteration of the local model training prior to performing the local SGD. Given that the former exploits the per-client gradient updates similar to the training data leakage at server aggregation, we focus on the latter and refer to such client-side attack as **training data leakage at client SGD**.

## 3.2 Training Data Privacy Leakage Formulation

Regardless of the specific attack implementation, the attack goal of training data privacy leakage is to reconstruct the private training data from the knowledge of gradients and federated learning model. Algorithm 1 gives a sketch of the training data privacy leakage from gradients. The attack configures and executes the reconstruction process in six steps. Concretely, (1) the adversary obtains the gradient update $\nabla_x f$ from the federated training process. (2) The attack algorithm $A : Z_x \to x_{rec}$ starts with a dummy seed $x_{rec}^0$ with the same resolution (or attribute structure for text) as the training data. (3) The dummy attack seed is fed into the client's local model. (4) The gradient $\nabla_{x_{rec}} f$ of the dummy attack seed is obtained by backpropagation. Since the local training update toward the ground-truth label of the training input data should be the most aggressive compared to other labels,

---

**Algorithm 1:** Gradient-based reconstruction attack

---

**Input**: training function $f: x \to Z_x$; $\nabla_x f$: stolen gradients; $INIT(x.init)$: attack
         initialization seed; $\mathbb{T}$: attack termination condition; $\eta'$ learning rate of attack
         optimization

    **// Attack procedure:**

1  $x_{rec}^0 \leftarrow INIT(x.init)$
2  $y_{rec} \leftarrow \arg\max_k (||\nabla_x f||_2)$
3  **for** $\tau$ *in* $\mathbb{T}$ **do**
4    |  $\nabla_{x_{rec}^\tau} f \leftarrow f(x_{rec}^\tau)$
5    |  $D^\tau \leftarrow ||\nabla_{x_{rec}^\tau} f - \nabla_x f||^2$
6    |  $x_{rec}^{\tau+1} \leftarrow x_{rec}^\tau - \eta' \frac{\partial D^\tau}{\partial x_{rec}^\tau}$
7  **end**
8  **Output:** reconstructed training data $x_{rec}$

---

the sign of gradient for the ground-truth label of the private training data will be different than other classes and its absolute value is usually the largest. Therefore, we can infer the label information from the class-wise gradient. (5) Given the gradient of the dummy data, the gradient loss is computed using a vector distance loss function, e.g., $L_2$, between the gradient $\nabla_{x_{rec}} f$ of the attack seed and the actual gradient $\nabla_x f$ from the client's local training. (6) The dummy attack seed is modified iteratively by the attack reconstruction learning algorithm. It aims to minimize the vector distance loss $D^\tau$ by a loss optimizer such that the gradients of the reconstructed seed $x_{rec}^i(t)$ at round $i$ will be closer to the actual gradient updates stolen from the client upon the completion (training data leakage at server aggregation) or during the local training (training data leakage at client SGD). When the $L_2$ distance between the gradients of the attack reconstructed data and the actual gradient from the private training data is minimized, the reconstructed attack data from the dummy seed converges to the private local training data, leading to the training data privacy leakage. This attack reconstruction iterates until it reaches the attack termination condition ($\tau$), typically defined by the maximum attack iteration or a specific loss threshold. If the reconstruction loss is smaller than the specified distance threshold, the training data leakage attack is considered successful.

Given the attack process, training data privacy leakage can be formulated as a reconstruction learning procedure $A : Z_x \to x_{rec}$, where $Z_x$ denotes the leaked gradient corresponding to private training data $x$ with the following attack objective:

$$\arg\min_{x_{rec}} ||\nabla_{x_{rec}} f - Z_x||_2. \tag{1}$$

The optimization goal is to iteratively modify $x_{rec}$ by minimizing the distance between the gradient of the reconstructed input $\nabla_{x_{rec}} f$ and the leaked gradient value $Z_x$: $||x - x_{rec}||_2 \approx 0$. Such that the reconstructed input $x_{rec}$ gradually becomes identifiably close to the private training data $x$ and eventually exposes the training example $x$ with high confidence as they become almost identical: $x_{rec} \approx x$. Figure 2 provides a visualization by three examples of Fashion-MNIST [88], CIFAR10 [36], and LFW [29] under training data leakage at client SGD.

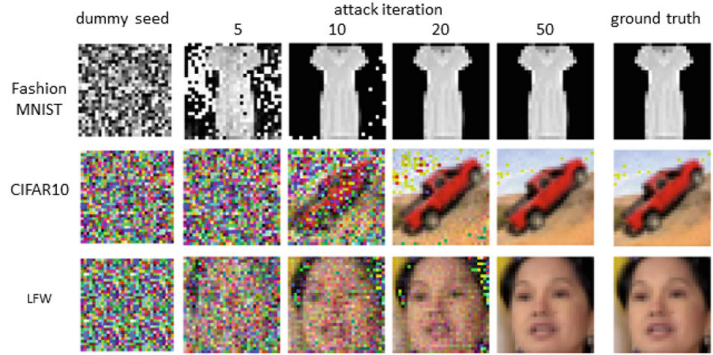

**Fig. 2** Reconstructive-based training data leakage attack at client SGD

For training data leakage at server aggregation, the leaked gradient of client $i$ is the accumulated result after the local training over the local training set $X$ at round $t$, denoted by $Z_X$. The reconstruction attack is to reverse engineer one of the private training examples in $X$ with $x_{rec}$.

$$\arg\min_{x_{rec}} ||\nabla_{x_{rec}} f - Z_X||_2. \tag{2}$$

Using different initial seeds, the same reconstruction inference attack algorithm can leak multiple private training data in $X$ such that $\exists_{x \in X} ||x - x_{rec}||_2 \approx 0$.

From the attack formulation and process, we make two interesting observations. *First*, multiple factors in the attack method could impact the attack performance of the training data privacy leakage, such as the dummy data initialization, the attack iteration termination condition, the selection of the gradient loss function, and the attack optimization method. For example, the bootstrapping initialization seeds significantly impact the attack stability, namely the reconstruction quality and convergence guarantee of the attack optimization, and attack cost, which is the number of attack iterations to succeed the reconstruction. Figure 3 provides a visualization of ten different initialization methods and their impact on the training data leakage attack in terms of reconstruction quality and convergence speed: random initialization seed, patterned initialization with 1/4 division and 1/16 division, patterned initialization with binary color of 0 and 1, patterned initialization with RGB colors, and initialization seed with another image from the same class. Figure 3 shows that all geometric initializations can outperform random initialization with faster attack convergence and better reconstruction quality.

*Second*, the configuration of some hyperparameters in federated learning may also impact the effectiveness and cost of the training data privacy leakage, including batch size and training data resolution. For example, the early gradient leakage attack algorithm in [101] uses separate weights and submodels for each training example (batch size of one) in order to show the reconstruction inference by reverse engineering and can succeed in the attack on the batch size of up to 8. The loss-function optimized attack algorithm in [21] shows the feasibility of an arbitrarily large batch of training data, e.g., a batch size of 100. By comparison, [83] show that when the input data examples in a batch belong to only one or two classes, which is often the case for mobile devices and the non-i.i.d. distribution of the training data [97], the training data leakage attacks can effectively reconstruct the training data of the entire batch, e.g., a batch size of 16 when the dataset has low interclass variation, e.g., face and digit recognition. Figure 4 shows the visualization of training data leakage at client SGD on the LFW dataset with four different batch sizes. We refer the readers to [83] for a comprehensive study on the influencing factors of training data privacy leakage. Choosing appropriate settings for these influencing factors can significantly impact attack effectiveness and cost.

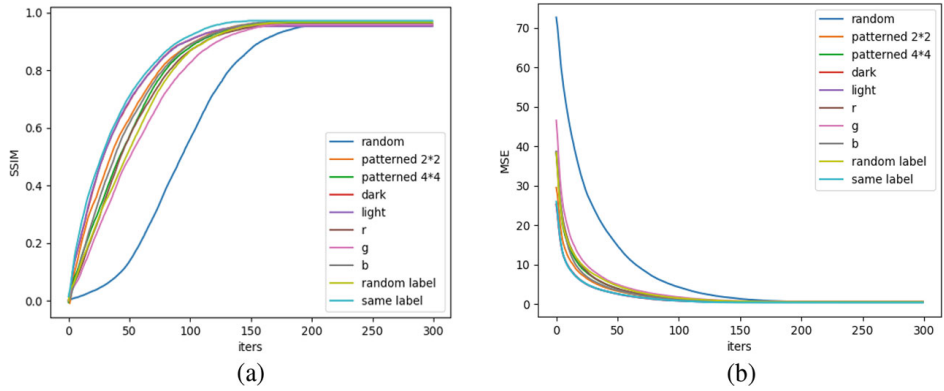

**Fig. 3** Attack convergence of CIFAR100 under different initialization seeds. (**a**) Structural similarity index measure (SSIM [80]). (**b**) Mean squared error

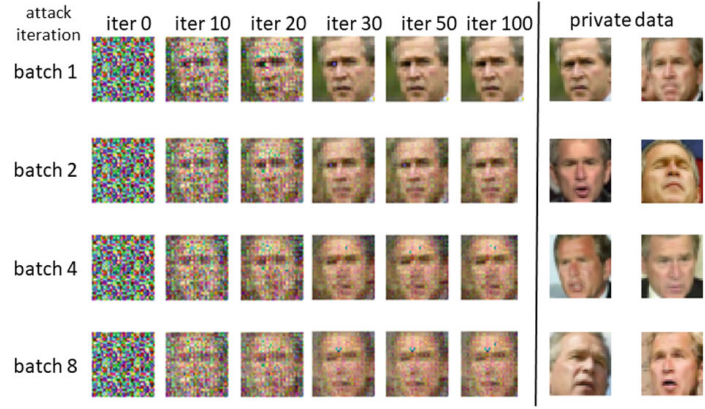

**Fig. 4** Effect of batch size in training data leakage at client SGD on LFW. Example from [81]

It is also worth noting to understand the difference of the attack reconstruction learning from the standard deep neural network training. In the latter, it takes as the training input both the fixed data–label pairs and the initialization of the learnable model parameters and iteratively updates the model parameters with gradients until the training converges. The learning process minimizes the loss with respect to the ground-truth labels. In contrast, training data leakage attacks perform reconstruction attacks by taking a dummy attack seed input, a fixed set of model parameters, such as the actual gradient updates of a client local training, and the gradient derived label as the reconstructed label $y_{rec}$, and its attack algorithm will iteratively reconstruct the local training data used to generate the gradient, $\nabla w_k(t)$, by updating the dummy synthesized seed data, following the attack iteration termination condition $\mathbb{T}$, denoted by $\{x_{rec}^0, x_{rec}^1, ...x_s^{\mathbb{T}}\} \in \mathbb{R}^d$, such that the loss between the gradient of the reconstructed data $x_{rec}^i$ and the actual gradient $\nabla w_k(t)$ is minimized. Here $x_{rec}^0$ denotes the initial dummy seed. If both the input query and the model are frozen, the federated model is used for label inference during deployment.

### 3.3 Observations on the Training Data Leakage Attacks

In this section, we speak out the untold truth about training data leakage attacks in terms of the access of the training model, the informative gradients in early training, and the effect of model perturbation with constant noise.

#### 3.3.1 Observation 1: Training Model Access

Our first observation on the training data privacy leakage is the implicit assumption that the adversary has the access to the local training model and can run the same training model for launching the iterative reconstruction-based inference attack. In other words, the adversaries have to access the training models used in federated learning to generate gradients from the initialization of dummy gradients during iterative attack optimization.

The necessity of access to the training model implies that the model leakage leads to the training data leakage. For the honest-but-curious server, the access to the training model is natural, and the server could collect gradient updates from every participating client, performing training data leakage attack at the FL server prior to aggregation. For adversary proxy at participating clients, even with the assumption that the adversary cannot access the encrypted data at rest, the training data privacy leakage remains feasible, assuming the attacker can gain access to the training model for reconstruction of private training data, for example, by running the same training model over the attack dummy seed (dummy initialization) against the stolen gradients.

For horizontal and vertical federated learning in which the clients do not share the gradient update with each other, training data leakage at client SGD can only reveal training data from the client where the adversary proxy resides. However, training data leakage at client SGD can disclose training data from those clients who share the gradients update to the adversary client in the peer-to-peer-based federated learning [81].

The observation also implies that training data leakage attack is rather difficult in the black-box setting. Suppose the adversary is unable to perform backpropagation on the training model. In that case, the attack optimization will not be able to update the dummy seed for its gradient converging to the stolen gradient. Although it is possible to find models whose gradients can approximate the gradient generated by the training model [4], the nonlinearity of deep learning models can lead to significant visual differences between the reconstructed instances and the private training data even when the approximated gradients are close to the stolen ones.

### 3.3.2    Observation 2: Impact of Attack Timing

Our second observation is that the stolen gradients at earlier training rounds of federated learning are more informative under the training data leakage attacks. The ability to reconstruct the private training data is much weaker on the gradient updates stolen from the later training rounds.

We attribute the phenomenon to the inherent logic of gradient descent. As federated learning progresses in rounds, the global model becomes more and more complex. The corresponding gradient generated on seen examples will demonstrate a decaying trend converging to 0. Figure 5 illustrates the effect of training data leakage attack after 1, 3, 5, 7, 9 local iterations. From this set of experiments, we observe that if the local model update can only be shared after the local training is performed over a certain number of iterations, then we can effectively reduce the probability of leaking the private training data at client even if the raw gradient updates are shared with the FL server.

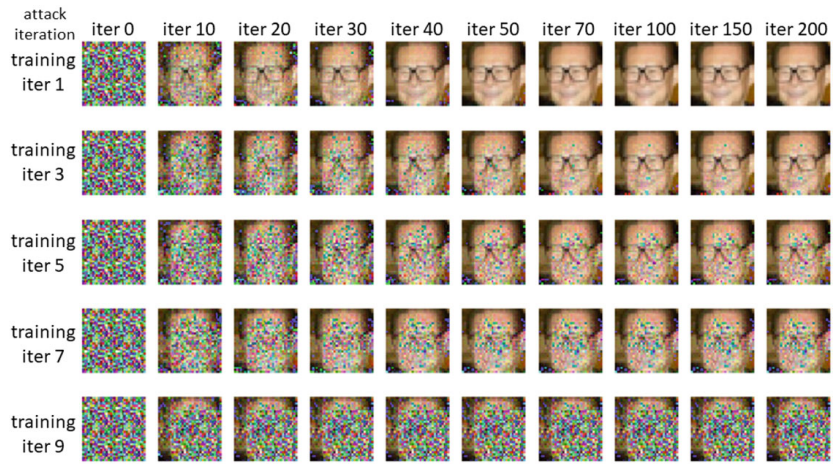

**Fig. 5**  Impact of training after several local training iterations. Example from [81]

### 3.3.3 Observation 3: Effect of Model Perturbation with Constant Noise

To protect gradient updates from training data leakage attacks, a common practice is refraining the participating clients from sharing their local model updates in raw format. Our third observation is that it is challenging to determine the proper amount of model perturbation to use. Existing model perturbation methods tend to use a constant perturbation strategy for ensuring training data privacy protection. Considering the different effectiveness of training data leakage at early and later rounds of federated learning, on one hand, using the constant amount of randomized noise for model perturbation may not be most effective to defend against training data leakage attacks. For example, at early rounds, such constant noise injection may be unnecessary, especially in later training rounds. On the other hand, by injecting excessive noise to a local model at later rounds may incur adverse effects on both accuracy and convergence of the global model. Therefore, adequate model perturbation should be employed to best balance the model performance and privacy protection.

## 3.4 Privacy Protection with Dynamic Perturbation

Existing model (gradient) perturbation methods for protecting training data privacy all adopt a straightforward data perturbation strategy by defining and adding a constant noise to all data at all time, such as gradient compression, randomized noise addition using Gaussian distribution, and differential privacy controlled noise injection. Consider conventional differential privacy (DP) parameters, such as using constant clipping bound to approximate sensitivity of the stochastic gradient descent (SGD) for Deep Neural Network (DNN) models using SGD optimizer [57]. Hence, a constant perturbation strategy is employed by most of the conventional DP algorithms. In the context of federated learning, to the best of our knowledge, [82, 84] are the first to inject dynamically generated randomized DP noise to sanitize the local model update prior to sharing with the federated aggregation server.

**Gradient compression** [41] sorts the gradients to be shared by a client and sends only the gradient coordinates whose magnitude is larger than a threshold. The approach removes the essential information needed for reconstruction [66, 83, 101].

**Gaussian noise addition** is another way to sanitize the raw gradients. A larger noise injection will alter the raw gradients more but may also hurt the model accuracy of federated learning.

Figure 6 illustrates gradient compression and Gaussian noise addition by example. We observe that under a low compression ratio of 10%, the gradient sanitization will have a low negative effect on the accuracy of federated but is vulnerable to training data leakage attacks. With a high compression ratio of 90%, we can gain training data privacy protection at the cost of decreased accuracy. Similarly, when choosing the small Gaussian variance threshold, the gradient sanitization fails to be resilient to training data leakage attacks. With a large Gaussian variance threshold, we gain leakage resilience at the cost of significant accuracy loss, from 0.695 with raw gradient to 0.344 under noisy gradient. We argue that (i) privacy protection with model perturbation may still intrude client privacy if insufficient perturbation is injected and (ii) it is hard to set a universal threshold for all models and all training tasks. Figure 6 shows the importance of choosing the appropriate model perturbation by balancing between leakage resilience and yet the minimal negative effect on the convergence and accuracy of federated learning.

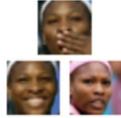

| | | gradient compression | | random noise | |
|---|---|---|---|---|---|
| raw data | non-private | 10% | 90% | 0.001 | 0.01 |
| test accuracy | 0.695 | 0.681 | 0.643 | 0.651 | 0.344 |
| leakage resilience | ✗ | ✓ | | ✓ | |

**Fig. 6** Gradient compression and Gaussian noise addition are hard to scale against training data privacy leakage

**Fixed Differential Privacy noise** is considered in conventional approaches to differentially private federated learning [23, 47, 84]. The noise is added either to the per-client model updates to protect against training data leakage at server aggregation [23, 47] or to the per-example local gradients to protect against both training data leakage at server aggregation and client SGD [84]. We refer the readers to the corresponding paper on the concrete implementation and differential privacy analysis of these differentially private federated learning approaches. Unlike Gaussian noise addition, differential privacy noise is controlled by differential privacy parameters $(\epsilon, \delta)$, and the $l_2$ norm of the gradient is capped by a predefined clipping bound for sensitivity control: $\mathcal{N}(0, \sigma^2 S^2 \mathbb{I})$ is injected, where the clipping bound $C$ approximates the sensitivity $S$, and $\sigma$ is the predefined fixed noise scale. $\mathbb{I}$ denotes the size of the noise reflecting the number of gradient coordinates.

Using a fixed clipping bound $C$ to define the sensitivity of gradient changes for all iterations can be problematic, especially for the later iterations of training since the fixed clipping bound $C$ to define sensitivity $S$ can be a very loose approximation of the actual $l_2$ sensitivity $S$: $S >> C$. With a fixed sensitivity $S$ and noise scale $\sigma$, the Gaussian noise with variance $\mathcal{N}(0, \sigma^2 S^2)$ will result in injecting a fixed amount of differential privacy noise throughout iterative federated learning. Injecting such excessively large constant noise to gradients in each iteration of the training may have a detrimental effect on the accuracy performance and slow down the convergence of training. Sadly, it does not gain any additional privacy protection because the accumulated privacy spending $\epsilon$ is only inversely correlated with $\sigma$ [82, 85].

Similar to the gradient compression and Gaussian noise addition, deciding how much perturbation to add for training data leakage prevention and model utility is difficult. Insufficient noise injected may maintain high model accuracy but fail to protect the model from training data privacy leakage. By comparison, excessive noise could prevent training data privacy leakage but at the cost of model performance.

Given that gradients at early training iterations tend to leak more information than gradients in the later stage of the training [83], it will be more effective to design a differential privacy algorithm with the amount of noise adaptive to the trend of gradient updates: injecting larger noise in early rounds and adding smaller noise to gradients in the later rounds during federated training. Given that the noise variance $\varsigma$ is the product of sensitivity $S$ and noise scale $\sigma$, several possible strategies can be promising, such as having the sensitivity calibrated to the $l_2$ norm of the gradients, or having a smoothly decaying noise scale such that the noise variance follows the trend of gradient updates across the entire training process.

**Dynamic Differential Privacy noise** considers dynamic differential privacy parameters. We introduce dynamic sensitivity $S$ defined by $l_2$-max of gradients and dynamic noise scale. The former strictly aligns to the gradient's $l_2$ norm and keeps track of the $l_2$ sensitivity of the local training model. Specifically, we promote to use the max $l_2$ norm of the per-example gradient in a batch as the sensitivity. By definition [16], the sensitivity of a differentially private function is defined as the maximum amount that the function value varies when a single input entry is changed. The definition indicates that the actual sensitivity of the function may vary for different input batches when performing local training at each client at each round $t$ of federated learning. Therefore, the $l_2$-max computed after clipping reflects more accurately the actual sensitivity of the local training function by following the sensitivity definition. Figure 7a shows the decaying trend of gradient updates in $l_2$ norm (blue curve), averaged over the participating clients at each round, as federated learning progresses in the number of rounds. This $l_2$-max sensitivity is dependent on the local training function. Hence, this $l_2$-max sensitivity is adaptive with respect to every local iteration, every client, and every round [82, 85].

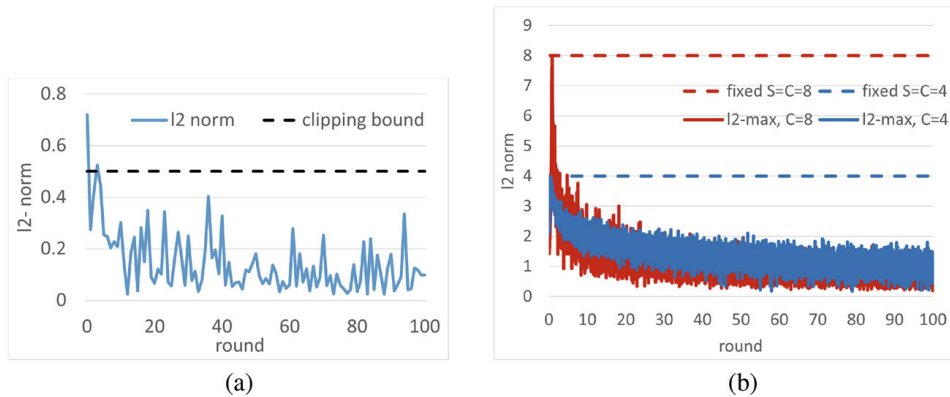

(a)                                                (b)

**Fig. 7** Decaying trend of the $l_2$ norm of gradient update in nonprivate federated learning and differentially private with fixed and dynamic differential privacy noise. Total clients $N = 100$ and participating clients $K_t/N = 10\%$ on Fashion-MNIST. (**a**) Vanilla federated learning. (**b**) Differentially private federated learning

Consider two scenarios: (i) When the $l_2$ norm of all per-example gradients in a batch is smaller than the predefined clipping bound $C$, then the clipping bound $C$ is undesirably a loose estimation of the sensitivity of training function under any given local iteration, client, and round. The max $l_2$ norm among the corresponding per-example gradients over the entire batch for iteration is, in fact, a tight estimation of sensitivity for noise injection. Instead if we define the sensitivity of the training function by the max $l_2$ norm among these per-example gradients in the batch, we will correct the problems in the above scenario. (ii) When any of the per-example gradients in a batch is larger than the clipping bound, the sensitivity of the training function is set to $C$. In summary, the $l_2$-max sensitivity will take whichever is smaller of the max $l_2$ norm and the clipping bound $C$. Figure 7b compares the fixed clipping-based sensitivity and using the $l_2$-max norm of the gradient to define the sensitivity $S$. When the $l_2$ norm of the per-example gradients in a batch is smaller than the fixed clipping bound $C$, using the clipping bound $C$ is a poor and undesirably loose approximation of the true $l_2$ sensitivity $S$ regardless of whether to set C=4 or C=8. Using fixed DP parameters to define gradient perturbation may lead to excessive noise injection and result in accuracy loss.

Dynamic noise scale with a decaying policy is an alternative approach to supporting dynamic differential privacy noise variance over the federated training process. This is because the differential privacy noise variance $\varsigma$ consists of both the sensitivity and noise scale. Dynamic noise scale can be implemented using a smooth decay function over the number of rounds in federated learning with different adaptive policies such as linear decay, staircase decay, exponential decay, and cyclic decay [85]. Each will progressively decrease the noise scale $\sigma$ as the number of rounds for federated learning increases. While we want to construct dynamic differential privacy noise, determining noise scale $\sigma_t$ will need to take the following three factors into consideration: (1) The starting noise scale $\sigma_0$ needs to be large enough to prevent gradient leakages. Note that general accuracy-driven privacy parameter search cannot always guarantee training data leakage resilience. Therefore, we select the privacy parameter settings proven empirically to be resilient [84] for the initial setting. (2) The ending noise scale $\sigma_T$ cannot be too small; otherwise the $\epsilon$ privacy spending would explode, resulting in poor differential privacy protection [81]. (3) The amount of noise injected is yet not too much to affect the desired accuracy performance of the global model.

Table 1 shows the comparison of fixed and dynamic model perturbation with differential privacy noise. We consider fixed differential privacy parameters: $C = 4$, $\sigma = 6$ as in [2, 84], and dynamic differential privacy parameters with $l_2$-max sensitivity $S$ and dynamic noise scale exponentially decaying from $\lceil \frac{C*\sigma}{S} \rceil$ to $\sigma_T = 3$. MSE measurement is the larger, the less similar between the reconstructed instances and private training data, with 0.4 as the threshold for successful reconstruction. The accuracy is measured at the round as in Table 2. By combining $l_2$-max sensitivity and dynamic noise scale, we are able to inject a larger noise at early rounds and a smaller noise at later rounds due to that the descending trend of $l_2$-max sensitivity results in the declining differential privacy noise variance as the training progresses.

**Table 1**  Comparison of fixed and dynamic model perturbation with differential privacy noise

|  |  | MNIST | Fashion-MNIST | CIFAR10 | LFW |
|---|---|---|---|---|---|
| No perturbation | Accuracy | 0.980 | 0.861 | 0.674 | 0.695 |
|  | MSE | 0.014 | 0.014 | 0.123 | 0.174 |
| Fixed perturbation | Accuracy | 0.956 | 0.826 | 0.633 | 0.649 |
|  | MSE | 4.95 | 4.92 | 2.77 | 2.79 |
| Dynamic perturbation | Accuracy | **0.977** | **0.854** | **0.642** | **0.683** |
|  | MSE | **5.03** | **5.06** | **2.89** | **2.86** |

**Table 2**  Benchmark datasets and parameters

|  | MNIST | Fashion-MNIST | CIFAR10 | LFW |
|---|---|---|---|---|
| # training data |  | 60000 | 50000 | 2267 |
| # validation data |  | 0000 | 10000 | 756 |
| # features |  | 28*28 | 32*32*3 | 32*32*3 |
| # classes |  | 10 | 10 | 62 |
| # data/client |  | 500 | 400 | 300 |
| # local iteration $L$ |  | 100 | 100 | 100 |
| Local batch size $B$ |  | 5 | 4 | 3 |
| # rounds $T$ |  | 100 | 100 | 60 |
| Vanilla accuracy | 0.980 | 0.861 | 0.674 | 0.695 |

**Fig. 8**  Convergence and $\epsilon$ spending accumulation of our methods with fixed and dynamic privacy parameters on MNIST

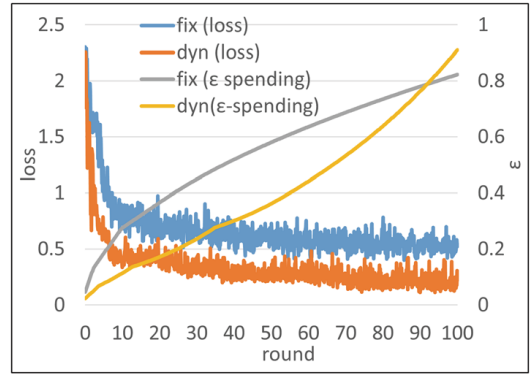

We also measure the impact of model perturbation by fixed and dynamic differential privacy noise on $\epsilon$ privacy spending and model convergence of federated learning. Following [2], we track $\epsilon$ spending using Rényi differential privacy [50] with a fixed $\delta = 1e - 5$. Figure 8 provides a visualization of comparing the loss over the global training rounds (x-axis) of federated learning for model perturbation by fixed differential privacy noise (blue) and dynamic differential privacy noise (orange), showing both guarantee the convergence, with fixed $\epsilon$ spending (gray curve) and dynamic $\epsilon$ spending (yellow curve).

## 3.5    Other Privacy Concerns in Federated Learning

### 3.5.1    Training Data Leakage Attacks Under Privacy-Enhancing Tools

The protection power of privacy-enhancing tools for securing data-in-use against privacy leakages varies depending on the attack surface. Secure multiparty computation (SMPC) is a cryptographic technique for enhancing privacy in multiparty communication and computation systems, such as securing per-client local model updates sharing with a remote and possibly untrusted aggregation server in federated learning systems [11, 52]. Hence, SMPC offers strong robustness against **training data leakage at server aggregation**, while having minimal impact on the accuracy of the global model. However, the main bottleneck of SMPC is the high communication cost. Also, SMPC may not secure the **training data leakage at client SGD** since the local SGD is performed on raw gradients of all examples in each minibatch per local iteration. Both homomorphic encryption (HE) and Trusted Execution Environment (TEE) are cryptographically capable of preventing training data inference attacks at client and at server in federated learning, as long as server and clients can support TEE or HE [51], respectively. For instance, in addition to running the aggregation server in TEE, each client can install TEE and ensure that both the local model training and local training data are hosted in the TEE enclave. However, enabling HE and TEE at both server and every client at global aggregation and every local SGD requires nontrivial cost, especially at edge clients with limited resources.

### 3.5.2    Other Privacy Intrusion Attacks Under Privacy-Enhancing Tools

Other known privacy intrusion attacks in federated learning include membership inference [32, 44, 54, 59, 64, 74], attribute inference [49], and model inversion attacks [19, 34, 65], which can be launched at both client and the federated server and cause more adverse and detrimental effects when combined with the gradient leakage attacks. Given that the discussion on the latter two attacks is rather limited, we will focus on the membership inference attacks.

Membership inference attack aims to infer whether a test data sample is a member of the training set based on the prediction result produced by a pretrained model during model deployment [64]. Membership inference attack on the trained federated learning model is the same as in the centralized setting. However, membership inference attack can also happen during the federated learning process as the training data are geographically distributed across a population of clients [74]. [49] introduce the first gradient-based membership inference attack in federated learning. The authors show that the nonzero gradients of the embedding layer of a recurrent neural networks model trained on text data can reveal which words are in the training batches of the honest participants. A possible explanation is that the embedding is updated only with the words that appear in the batch, and the gradients of the other words are zeros. [54] temper with the federated training process and intentionally update the local model parameters to increase the loss on the target data record. If the target data record is a member of the training set, applying gradient ascent on the record will trigger the model to minimize the loss of this record by gradient descent, whose sharpness and magnitude are much higher than performing gradient ascent on data records that are not members of the training set. Different proposals have been put forward for enhancing robustness against membership inference, including differential privacy [74], prediction confidence masking [14, 26, 35], regularization [43, 53], dropout [37, 59], model compression [78], knowledge distillation [58, 62] that have been proposed to alleviate the membership inference attack. However, these techniques can provide only limited robustness against the membership inference, for example, by lowering its attack success rate by 20% ∼ 30%, and none can eliminate the privacy threat completely or at a high defense success rate [73]. Also, given that membership inference attacks during the federated training process, privacy-enhancing techniques such as HE and TEE cannot protect the private training data from the attack until it is inside the enclave.

## 4 Data Poisoning and Security Assurance

### 4.1 Threat Model

Poisoning attacks during the federated training assume malicious clients and can be performed on data or model. Data poisoning attack occurs during local data collection and has two types: 1) clean label [61] and 2) dirty label [25]. Clean-label attacks inject training examples that are cleanly labeled by a certified authority. Imperceptible adversarial watermarks are injected to the clean input to form a poisoning instance with a clean label but simultaneously minimize the $l_2$ distance of the input to the target instance. In contrast, dirty-label poisoning deletes, inserts, or replaces training examples with the desired target label into the training set. One example of dirty-label poisoning attack is backdoor poisoning [25], in which the adversary inserts small regions of the original training data and modifies the label as the desired target class to embed the trigger into the model. In this way, the unaltered input will not be affected, and the input with the trigger will behave according to the adversary's objective [7, 67, 70, 76]. Another example is the label-flipping attack [9, 20, 71], which flips some source victim class to another designated target class, while the features of the data are kept unchanged. Model poisoning attack happens during the local model training process, aiming to poison local model updates before sending them to the server. Since data poisoning attacks eventually change a subset of updates sent to the model at any given round, model poisoning is believed to subsume data poisoning in federated learning settings [8].

Depending on the attacker's objective, poisoning attacks can be either: a) denial-of-service random attacks or b) stealthy targeted attacks. The former aims to reduce the accuracy of the federated learning model, whereas the latter seeks to degrade the performance of a particular source class (victim) or induce the federated learning model to output the target label specified by the adversary while keeping high test accuracy on the rest of the classes. Targeted attack is considered more difficult than random attacks as the attacker has a specific goal to achieve but is more motivated since the attacker can manipulate the model for its adverse goal. Accordingly, the main focus of our study is on the targeted data poisoning attacks: targeted dirty-label poisoning, backdoor attacks, and clean-label attacks. These attacks assume that each malicious client can only manipulate the training data $X_i$ with auxiliary information such as the target label on their own device but cannot access or manipulate other participants' data. These attacks corrupt training data with different tactics but remain the learning procedure, e.g., SGD, loss function, or server aggregation unaltered. These attacks are not specific to any deep neural network architecture, loss function, or optimization function. Also, these attacks are stealthy as they succeed in dropping the prediction accuracy of the manipulated input, and yet the poisoning attack has little negative impact on the accuracy of the rest of the queries.

### 4.2 Training Data Poisoning Attack Formulation

#### 4.2.1 Targeted Dirty-Label Poisoning

Targeted dirty-label poisoning corrupts training data with label change [71]. Let $F(x)$ denote the global model being trained in federated learning, $f_i(x)$ be the local model of client $i$, and $(x, y)$ denote the raw data and its ground-truth label in the training set of client $i$. The objective of the poisoning attack $\rho$ is to replace the ground-truth label $y$ with $y'$ to mislead the joint training so that the global model produced by federated learning can be fooled. The global model will mispredict examples of source class $y$ to target class $y'$ with high confidence, formally:

$$\rho : \rho(x, y) = (x, y')$$

$$s.t. \quad f_i(x) = y', \ y' \neq y, \ \max[F(x) = y'].$$

The objective of the targeted dirty-label poisoning attack is to maximize the chance of the global model F(x) to misclassify the test examples of the source class, by poisoning the training data of the source class on those of compromised clients.

#### 4.2.2 Backdoor Poisoning

Compared to the targeted dirty-label poisoning, backdoor attackers corrupt training data by injecting triggers such that input queries with the trigger will misbehave, while the input queries without the trigger will act normally [7, 67, 76, 89]. With $\delta x$ as the trigger and $x' = x + \delta x$, we can formulate backdoor poisoning as

$$\rho : \rho(x, y) = (x', y')$$

$$s.t. \quad f_i(x') = y', \ y' \neq y, \ \max[F(x') = y'].$$

The objective of the backdoor poisoning is to maximize the chance of the global model F(x) to misclassify the test examples with the trigger, by inserting triggers to the training data on those compromised clients.

### 4.2.3   Clean-Label Poisoning

Unlike dirty-label and backdoor poisoning, clean-label poisoning attacks add another layer of inputs to the original inputs such that injected features overtake the original features [22, 31, 61, 99]. Clean-label poisoning uses the gradient-based procedure to optimize how the training examples are poisoned to prevent detection. Let $x^*$ be the input from the target class and $x' = x + \beta x^*$, where $\beta$ is commonly set smaller than 0.5, and we can formulate clean-label poisoning as

$$\rho : \rho(x, y) = (x', y)$$

$$s.t. \quad f_i(x') = y', \ y' \neq y, \ \max[F(x') = y'].$$

The objective of the clean-label poisoning is to maximize the chance of the global model F(x) to misclassify the test examples embedded with inputs from another class. The resulting model will make decisions based on the injected features on the top instead of the original features.

   To increase poisoning data participation for more severe poisoning effect in federated learning, one straightforward approach is to engage with more compromised clients. Namely, the percentage ($\lambda$) of compromised clients is large. However, poisoning attackers typically assume a percentage of comprised clients, e.g., 5%, 10%, or 20% of the total $N$ participating clients to avoid outlier detection. In this case, the number of poisoned local training data examples is limited. To make effective poisoning attacks, strategic adversaries may purposely increase the participation of these compromised clients [71]. For example, some distributed learning services require a stable power supply and fast WiFi connectivity [10]. Attackers can thus make themselves always available at times when insufficient honest participants are available, so that malicious clients have a higher probability of being selected by the federated learning server during each round of the joint training. In other words, while the percentage ($\lambda$) of comprised clients is small, the $\alpha$ chance that the gradient update collected by the server is from a malicious client is large.

## *4.3 Observations on the Training Data Poisoning Attacks*

In this section, we first uncover the unspoken fact of training data poisoning attacks in terms of model access, attack timing, and other key factors that impact on poisoning effectiveness. Then we discuss the myth and the effect of employing the DP model perturbation as a method to mitigate the training data poisoning attacks.

### 4.3.1   Observation 1: Training Data Access

Based on our extensive experiments on substantial collection of existing data poisoning attack methods, we observe that to launch a data poisoning attack, be it dirty label or clean label, the baseline assumption is that the adversary has the access to the training data hosted privately at local clients. This indicates that the data poisoning attacks do not need to directly modify the model, as suggested in [8], and instead the adversary is assumed to have access to the local training data on the compromised client and hence can access the training data at run time, even though the training data at rest is encrypted. As a result, adversaries can directly and strategically poison the ground-truth data, such as flipping the label or adding backdoor triggers only to the training examples of some victim class, while keeping the remaining of the training data untouched [81]. In most of the data poisoning attacks, the adversary may have zero knowledge about the DNN model structure and its hyperparameter settings when the model trojan attack is simply to poison the target data of victim class by flipping the ground-truth label or injecting backdoor trigger to misguide the prediction input query into the targeted poisoning trap, such as changing the prediction from correct source class to an attack target class through targeted poisoning using dirty label or backdoor trigger. In the backdoor trigger case, the same backdoor trigger (patch) once planned into the prediction query input, it will result in misguiding a well-trained DNN model to deliver a wrong prediction (either targeted or untargeted attack).

It is worth to note that most of the data poisoning attacks are targeted. First, attackers only selectively poison some or all training data of a chosen victim class while keeping the rest of the classes untouched. To perform targeted poisoning by either injecting backdoor or modifying ground truth, the attackers are assumed to have the access to the targeted training data and can read and manipulate these training data. Hence, encryption at rest cannot prevent such poisoning risks. However, DNN model training directly on encrypted data is still in its infancy and remains an important research problem for AI security, especially in federated learning environments.

### 4.3.2   Observation 2: Impact of Attack Timing

Our second observation is that while data poisoning attacks can occur at any iterative round during the entire course of federated learning, and last for an arbitrary

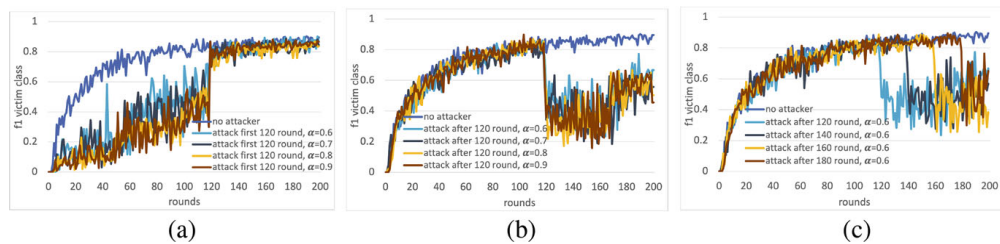

**Fig. 9** Different attack timing on CIFAR10 by poisoning the victim class (class 1) at availability $\alpha = 0.6, 0.7, 0.8, 0.9$ and $\lambda = 10$. Results from [81]. (**a**) Poisoning first 120 rounds. (**b**) Poisoning last 60 rounds. (**c**) Attack timing in later rounds

number of rounds, the poisoning attacks are more effective at the later stage of training compared to only performing poisoning in the early stage and stopping at the midway [71]. We attribute the phenomenon to the catastrophic forgetting [24] characteristics of deep learning models. When trained on one task, then trained on a second task, deep learning models "forget" how to perform the first task. Figure 9a demonstrates the attack effect of the early attackers who inject data poisoning for the first 120 rounds for CIFAR10. Percentage $\lambda$ of comprised clients is set to 10%, and the $\alpha$ chance that the gradient update collected by the server is from a malicious client is set to 60%, 70%, 80%, 90%. The results show that if the poisoning attacker only gets involved at the early stage of training and then leaves for good, later rounds of clean training would correct the altered poisoning effect. By comparison, Figure 9b shows the results of late-stage attacks. The late-round attack is more effective in degrading the performance of the victim class on the model to be published at round 200 for CIFAR10.

There are some other worth noting empirical observations. For example, it usually takes several rounds for the poisoning attack to be effective [81]. If the attacker fails to perform sufficient rounds of poisoning attacks on a compromised client, the poisoning effect on the local model update shared by this client to the FL serve may not effectively hurt the aggregated global model, which is learned from multiple rounds of distributed learning and multiple and possibly diverse participating clients in each round. Therefore, engaging in the poisoning activity but stopping too early or launching poisoning attack too late will both result in a poor poisoning attack effect. Figure 9c shows that the repairing power of the benign clients is not very strong, and the data poisoning would remain effective for longer rounds, e.g., 30 rounds–50 rounds.

### 4.3.3 Observation 3: Model Perturbation with Constant Amount of Noise

There are several threads of efforts to mitigate risks of training data poisoning attacks. One threat of existing solutions is to train a global model using a differentially private federated learning approach. This requires to add a constant amount of noise to local model/gradient update at each round. As a result, the use of perturbed local model update will cancel some adverse effects of data poisoning attack for both the local gradients produced by compromised clients and the global model, which is aggregated from noisy local model updates. To constrain the negative effect of gradient perturbation performed at the honest/benign clients, we need to determine the amount of noise to be used for model perturbation is not too much in order to maintain the acceptable accuracy of the global model, and at the same time, we need also to ensure that the amount of noise should be sufficient to mitigate/cancel the effect of data poisoning. Seeking a good balance between poisoning resilience and model accuracy is known to be a nontrivial technical challenge.

Given that most existing model perturbation approaches [7, 45, 55, 67, 76] use the constant amount of randomized noises, such as model perturbation using the conventional differential privacy controlled noise. However, we observe from extensive empirical measurements that it is critical and yet challenging to determine the proper amount of model perturbation to use at different rounds of federated learning. First, the early rounds usually produce larger model gradient updates compared to later rounds. By using a constant amount of random noise for model perturbation, we may add too much (excessive) noise in later rounds, which can negatively affect the accuracy and convergence of the global model because gradients will become smaller as the federated training rounds are progressing. Furthermore, the poisoning effects at early stage of the federated training tend to be less effective compared to poisoning performed only in the later rounds of federated learning. Hence, employing constant noise across all rounds of the federated learning is not optimal for maintaining good performance of the global model. This is especially true when the model perturbation is employed solely for mitigating data poisoning effect.

To the best of our knowledge, there are little efforts to date that set forth for developing model perturbation solutions for safeguarding federate learning against both training data privacy leakage and training data poisoning threat.

## 4.4 Boosting Poisoning Resilience with Dynamic Model Perturbation

Bearing the above discussion and analysis in mind, in this section we discuss opportunities of employing dynamic model perturbation strategies. Unlike existing model perturbation methods with a constant perturbation strategy, the dynamic model perturbation methods will seek to find the appropriate model perturbation by balancing between data poisoning mitigation and the minimal negative effect on the convergence and accuracy of federated learning. In some sense, the dynamic model perturbation for poisoning resilience shares some analogy to federated learning with differential privacy. But they differ in at least one fundamental perspective. Conventional differential privacy defines the constant amount but randomized noise addition with the goal of ensuring that the noise is large enough under acceptable model accuracy loss (controlled by a user-defined privacy budget). Hence, the level of privacy protection by differential privacy is defined by this privacy budget. However, for poisoning resilient model perturbation we need to define the amount of noise to add based on the poisoning mitigation effectiveness such that we can remove or eliminate the poisoning effect while maintaining the acceptable model accuracy loss.

**Fig. 10** $L_2$ norm of benign and poisoned gradient update. Total clients N=100 and participating clients $K_t/N =$ 10 % on Fashion-MNIST. Results from [81]

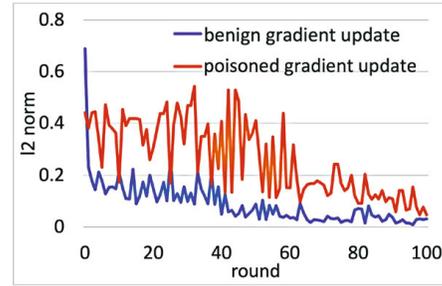

**Fig. 11** Decoupling of benign (yellow dot) and poisoned (blue cross) gradient update under data poisoning attack. Measured with $\lambda =$ 10% when flipping source class 1 to target class 9 of CIFAR10

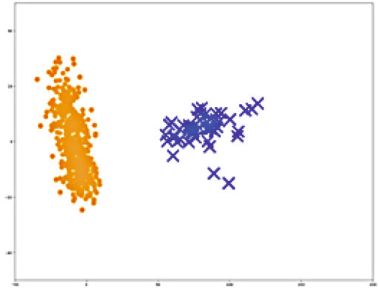

Figure 10 shows the $l_2$ norm of the gradient update for both benign and poisoned settings of federated learning. We argue that the more effective poisoning effect at the later stage of training results in the larger gradients from unseen/less seen poisoned update, while the benign gradient update converges to 0 due to gradient descent.

To demonstrate the impact of model perturbation on the poisoning effect, we resort to the gradient decoupling phenomenon [81] on the eigenvalues of the covariance in the gradient update shared from the client to the server. Specifically, the distribution of benign gradients from honest clients can be separable from the distribution of poisoned gradients from compromised clients by performing Principal Component Analysis (PCA) or clustering to the gradient updates at the federated server [15, 71], as shown in Fig. 11. Figure 12a shows that model perturbation with a small constant differential privacy noise has little impact on the gradient decoupling with $\lambda = 10\%$. Figure 12b shows the measurement results for a large constant differential privacy noise. The noisy gradients can cancel the poisoning effect when only a small percentage of malicious clients is present. We can interpret this phenomenon based on the output stability of DP [48], which states that DP noise perturbation is an $e^{\epsilon} - 1$ dominating strategy slightly deviated from the mainstream direction of the gradient update. When the amount of malicious clients is limited, differential privacy noise would bring the poisoned gradient direction

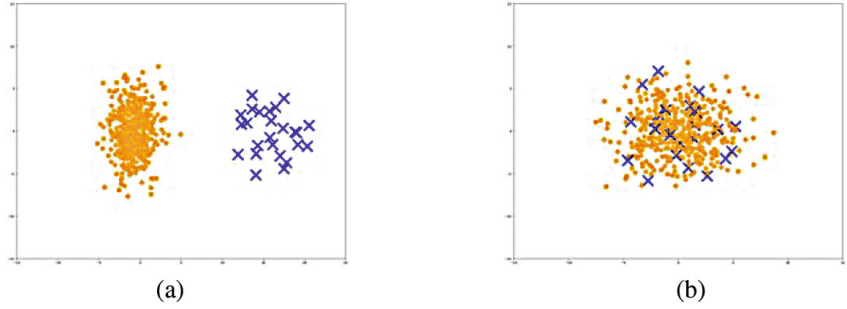

**Fig. 12** Gradient decoupling effect under differential privacy noise, measured in CIFAR10. (**a**) $C = 0.1$, $\sigma = 0.1$. (**b**) $C = 0.5$, $\sigma = 2$

back to the right track. However, when the percentage of the malicious clients is large, e.g., $\lambda > 50\%$, there is a high probability that the majority of the gradient updates on the source class at some round(s) may be dominated by poisoned contributions from malicious clients.

With the above empirical observations in mind, we conjecture that using the dynamic model perturbation designed by our dynamic differential privacy optimization outlined in Section 1.3.4 can be a viable solution [81]. Next, we show how dynamic noise can be significantly more effective in mitigating data poisoning attack than using the constant amount of noise as done in conventional differential privacy methods [1]. Recall Section 1.3.4, we use the $l_2$-max sensitivity instead of constant clipping bound to define the amount of random noise to be added for model perturbation, and this allows dynamic DP noise to be computed based on the gradient fluctuation in each round of federated learning. With a proper setting of initial noise scale and corresponding noise variance, we measure the impact of using dynamic DP-controlled noise in mitigating poisoning attacks and report the result in Table 3. We make three observations: (1) With sufficiently large noise, dynamic model perturbation is not only leakage-resilient (shown in Fig. 13) but also offers good poisoning resilience under m = 5% and m = 10%. (2) With the initial noise variance $S_{dyn} * \sigma_0 = 5$, dynamic differential privacy noise leverages a decaying noise variance that is large enough at early rounds for leakage resilience and decreases by following the declining trend of $l_2$-max sensitivity as the number of rounds increases. The early poisoning resilience comes from the output stability that cancels the effect of the poisoned gradient. (3) At the later stage, the added differential privacy noise for leakage resilience becomes smaller and may no longer effectively cancel out the effect of the poisoned gradient. Combined with the PCA-based gradient outlier removal mitigation, the poisoning resilience can be further improved by 5–10% for all three datasets.

By analyzing the effectiveness of dynamic perturbations against both training data poisoning and training data leakage attacks, we make the following remarks for developing security strategies in federated learning to simultaneously mitigate both security and privacy threats:

**Table 3** Poisoning resilience of dynamic differential privacy noise measured in micro f1 score

| sample | target | m | No perturbation | | Dynamic perturbation | | Dynamic perturbation + outlier removal | |
|---|---|---|---|---|---|---|---|---|
| | | | victim class | rest classes | victim class | rest classes | victim class | rest classes |
| 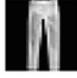 | ankle | benign | 97.0 % | 88.4 % | 95.8 % | 87.7 % | 96.8 % | 88.1 % |
| | | 5 % | 82.2 % | 88.4 % | **91.2 %** | 87.7 % | **96.8 %** | 88.1 % |
| | | 10 % | 44.9 % | 88.3 % | **85.3 %** | 87.7 % | **95.6 %** | 88.1 % |
| 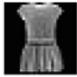 | shirt | benign | 92.5 % | 88.9 % | 90.5 % | 86.5 % | 91.4 % | 68.0 % |
| | | 5 % | 76.0 % | 88.8 % | **86.7 %** | 86.4 % | **88.2 %** | 68.0 % |
| | | 10 % | 51.6 % | 88.8 % | **82.5 %** | 86.4 % | **86.9 %** | 68.0 % |
| 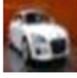 | truck | benign | 88.1 % | 72.6 % | 85.9 % | 68.0 % | 87.4 % | 68.0 % |
| | | 5 % | 75.6 % | 72.7 % | **82.5 %** | 68.0 % | **87.3 %** | 68.0 % |
| | | 10 % | 50.3 % | 72.7 % | **78.7 %** | 68.0 % | **87.1 %** | 68.0 % |
| 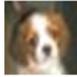 | cat | benign | 78.4 % | 73.8 % | 74.8 % | 70.2 % | 77.1 % | 71.9 % |
| | | 5 % | 66.5 % | 73.8 % | **72.1 %** | 70.2 % | **76.0 %** | 71.9 % |
| | | 10 % | 40.3 % | 73.8 % | **69.6 %** | 70.7 % | **73.9 %** | 71.9 % |
| 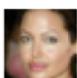 | Jennifer Aniston | benign | 68.7 % | 69.6 % | 67.2 % | 67.5 % | 68.3 % | 69.1 % |
| | | 5 % | 59.1 % | 69.6 % | **64.9 %** | 67.5 % | **68.2 %** | 69.1 % |
| | | 10 % | 46.8 % | 69.6 % | **60.5 %** | 67.5 % | **67.8 %** | 69.1 % |
| 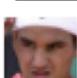 | Tiger Woods | benign | 70.6 % | 69.4 % | 67.9 % | 67.1 % | 68.6 % | 68.3 % |
| | | 5 % | 62.3 % | 69.4 % | **65.1 %** | 67.1 % | **68.4 %** | 68.3 % |
| | | 10 % | 51.1 % | 69.3 % | **60.9 %** | 67.0 % | **67.7 %** | 68.2 % |

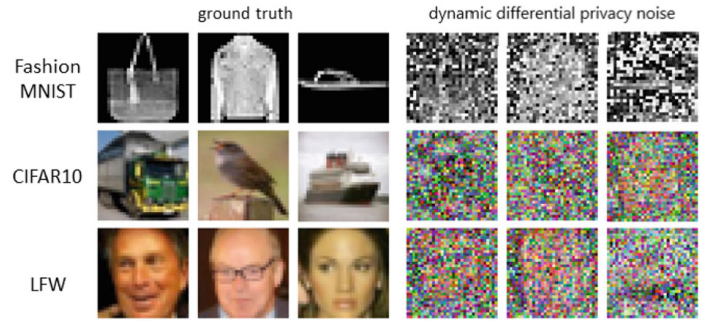

**Fig. 13** Leakage resilience of dynamic differential privacy noise

- **Remark 1.** From Fig. 10, we make two observations: First, the gradient effect of poisoning attacks remains similar across all rounds of federated learning, regardless of the attack timing of data poisoning. Second, the poisoned gradients tend to be consistently larger than the benign gradients. This is one of the main reasons that poisoning attack in the later half of the federated learning rounds will have more detrimental effect on the victim class, compared to the poisoning attacks performed only in the early rounds of federated learning (recall Fig. 9).

- **Remark 2.** Although gradient perturbation may help mitigate the poisoning effect to some extent, it remains an open research question regarding how to determine the right amount of model perturbation at each round of federated learning. This is because on one hand we need to perturb the client model update with sufficiently large noise to cancel the negative effect of poisoning, and on the other hand, we need to ensure the amount of noise used for model perturbation is just enough and not too large in order to preserve the accuracy of global model. Table 3 shows that while noise injection can partially remove the poisoning effect, the accuracy of the nonvictim classes drops as well, even with dynamic model perturbation method.

- **Remark 3.** The model perturbation method for poisoning mitigation must assume that the percentage of malicious clients is small [45, 55]. This is because the protection power of differential privacy controlled noise is an $e^\epsilon - 1$ dominating strategy slightly deviated from the mainstream direction of the gradient update.

We argue that the security protection techniques for federated learning should bear the above analysis and observations into consideration when determining the right amount of noises to be used by the model perturbation. Strategic model perturbation approaches, such as selective noise injection only on the largest gradients, are one possibility to explore.

## *4.5 Categorization of Poisoning Mitigation Techniques*

### 4.5.1 Server-Side Mitigation Techniques

Existing defense solutions against poisoning attacks rely on the assumption that the federated server in distributed learning is trusted. Hence, the primary research efforts are dedicated to detecting anomalies by separating poisoned and nonpoisoned contributions. Most existing poisoning defense solutions are based on the detection of poisoned local model updates sent from the compromised clients.

**Spatial Signature-Based Techniques** Tolpegin et al. [71] propose to apply PCA on the local model updates collected over multiple rounds for each class and produce two distinct gradient clusters for each poisoned source class. One corresponds to benign local model updates from honest clients, and the other corresponds to the poisoned local model updates from compromised clients. Based on the assumption that only a small percentage of participating clients are compromised, it considers the smallest cluster of the two will be the poisoned gradients from compromised clients. [39] score model updates from each remote client by measuring the relative distribution over their neighbors using a kernel density estimation method and distinguishing malicious and clean updates with a statistical threshold. [72] perform spectral analysis with SVD to generate two clusters for backdoor poisoning attacks. [27] utilize robust covariance estimation to amplify the spectral signature of corrupted data for detection. [38] conduct spectral anomaly detection using variational autoencoder with dynamic thresholds. [69] propose to decompose the input image into its identity part and variation part to perform statistical analysis on the distribution of the variation and utilize a likelihood-ratio test to analyze the representations in each class to detect and remove the backdoor trigger.

**Spatial-Temporal Signature-Based Techniques** STDLens [15] is the first work to identify the problem of treating the smaller cluster of the two as the poisoning gradients (Trojan attacked local model updates). In addition to spatial signature generated with PCA and k-means clustering over the local model updates collected over multiple rounds for each class, STDLens introduces the temporal signature as the second step dedicated to identify which of the two gradient clusters is the poisoned gradients. Instead of removing the entire cluster of poisoning gradients, STDLens identifies another technically challenging case where the PCA with K-means fails to partition the gradients of a class from the participating clients of a given round into two cleanly separated clusters. This is because simply removing the cluster of poisoned gradients may result in removing benign gradients and honest clients. STDLens addresses the problem of two overlapping clusters by employing the $\lambda$ density analysis to filter out the uncertainty region around the overlapping of the two clusters prior to executing the removal of poisoning gradients and the corresponding clients who shared the poisoning gradients with the federated server. It is worth noting that this chapter is the first to introduce three types of poisoning attacks to DNN object detection models: poisoning object existence, poisoning object bounding box by shuffling them over different locations of the input image, and poisoning the label of the victim class.

**Meta-Learning-Based Techniques** Xu et al. [90] train a meta-classifier that predicts whether a given target model is Trojaned due to data poisoning. Specifically, the authors introduce a technique called jumbo learning that samples a set of Trojaned models following a general distribution and offline learn a Generative Adversarial Network (GAN)-based meta-classifier to determine whether a local model is Trojaned. During online Trojan detection, the meta-learning method will run at the server and evaluate every local model received by the server and reject those models that are detected as Trojaned models before performing global model aggregation.

**Server-Side Validation** Server-side validation either assumes that the federated server has a clean validation dataset with benign (untainted) ground-truth labels or assumes that the clients can cross-validate each other with no collusion. The validation can be done every round or on selected rounds. [56] train a k-Nearest Neighbors (kNN)-based distinction classifier with a validation dataset to filter out the poisoned samples. [96] require the server to send local model updates from some clients to other clients for cross-checking. [13] require the service provider to collect a clean small training dataset and bootstrap the trust score for each client. A local model update has a lower trust score if its direction deviates more from the direction of the server model update. Then, the server normalizes the magnitudes of the local model updates such that they lie in the same hyper-sphere as the server model update in the vector space, thus limiting the impact of malicious local model updates with large magnitudes. CONTRA [5] implement a cosine-similarity-based measure to determine the credibility of local model parameters in each round and a reputation scheme to dynamically promote or penalize individual clients based on their per-round and historical contributions to the global model. Li et al. [40] find that the models can learn backdoored data much faster than learning with clean data. Therefore, they introduce a gradient ascent-based anti-backdoor mechanism into the standard training to help isolate low-loss backdoor examples in early training and unlearn the backdoor correlation.

### 4.5.2 Neural Network Cleansing Techniques.

An alternative countermeasure against poisoning attacks is to perform neural network cleansing, which sanitizes the model or its input to remove the poisoning effect.

**Input Sanitization** For input sanitization, one example is to regularize the class boundaries on the convex combinations of training data points [12]. By this means, the small nonconvex regions are removed, which causes a poisoned data instance being surrounded by (nonpoisoned) instances with different labels, and thereby mitigating the effect of poisoning. Another study [75] finds that for an infected model, it requires much smaller modifications on the input to cause misclassification into the target label than into other uninfected labels. Therefore, they can iterate through all labels of the model and determine if any label requires significantly a smaller amount of modification to achieve misclassification. If a backdoor is identified in the model, the proposed method can produce the trigger responsible for the backdoor. Accordingly, a proactive filter can be built to detect and filter out all adversarial inputs that activate backdoor-related neurons.

**Model Sanitization** In addition to model perturbation by adding randomized noises, other methods for model sanitization share similar objectives, which is to prune the dormant neurons to weaken the poisoning impact [60]. Li et al. [40] report that the models can learn backdoored data much faster than learning with clean data. Therefore, they introduce a gradient ascent-based anti-backdoor mechanism into the standard training to help isolate low-loss backdoor examples in early training and unlearn the backdoor correlation. Wu and Wang [87] show that model sanitization can also be done after the model has been fully trained and poisoned. Based on the observation that the poisoned neurons are easier to collapse after adding adversarial noise on them, they formulate a min-max problem to alternatively optimize the adversarial noise, which serves to expose the poisoned neurons, and the mask, which serves to prune out the poisoned neurons. By pruning out the poisoned neurons as indicated by the mask, the model is fully recovered from the backdoor behavior. CLP [98] utilizes a similar idea of pruning, but they utilize a different criterion—channel Lipschitz constant to identify the poisoned channel—and similarly remove the suspected channels afterward.

**Model Sanitization in Federated Learning Context** We test CLP pruning [98] on a poisoned model trained on centralized/federated learning procedure [30], whose results are available in Fig. 14. As shown in the left figure, CLP pruning may drastically decrease the benign accuracy when adopting a large pruning ratio,

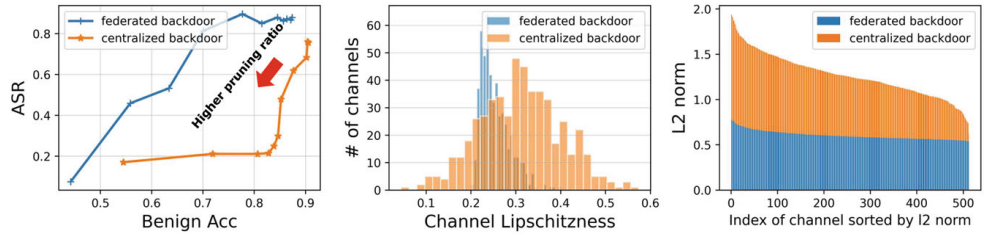

**Fig. 14** Properties of two models trained with centralized backdoor and federated backdoor. Left: ASR and benign accuracy with CLP defense. Middle: Channel Lipschitz of the last convolutional layer of two models. Right: L2 norm of last convolutional layer of two models

which is necessary to lower Attack Success Ratio (ASR) to a satisfied number. We also see from the middle/right figure that for a federated backdoored model, the Lipschitz constant and the L2 norm of different channels (parameters) do not show substantially difference, which make it harder to identify the poisoned parameters in a statistical way. This indicates that pure pruning defense may not work well in federated learning context, and extra counter-measurement needs to be taken in the training phase (e.g., isolation subspace training in [30]).

## 5   Other Risk Factors in Federated Learning

While most discussions on the security threats of federated learning today focus on training data privacy intrusion and training data poisoning attacks, the distributed nature of federated learning introduces additional security challenges. The lack of centralized control makes it difficult to enforce stringent security measures on each client (edge device). This opens doors to malicious participants to manipulate and compromise the federated learning process and outcomes.

### 5.1   Data Skewness and Biases

Skewness measures the distortion of symmetric distribution in a dataset. Skewness is a significant issue in federated learning because the distribution of data across different devices or clients varies significantly. This imbalance in data distribution can lead to biased and suboptimal model updates. For example, certain devices may contribute disproportionately more or less data than others. Such skewed data can result in models that are biased toward data-rich clients and perform poorly on data-poor clients, ultimately compromising the overall performance and generalization of the federated model. In the meantime, the disparity of the majority and minority of classes in a skewed data distribution can be amplified by differential privacy noise [6]. Addressing data skewness in federated learning is essential to ensure a fair representation of all clients' data and to improve the collective model's accuracy and robustness. Strategies like balanced sampling, loss reweighting, and gradient tuning [63, 77] are among the approaches to tackle this challenge and achieve more balanced and reliable federated learning outcomes.

## *5.2   Misinformation*

The issue of misinformation is another significant concern in federated learning, especially in scenarios where data is sourced from multiple devices or clients. Since federated learning involves training a global model using decentralized data, there is a risk of including misinformation or malicious data from individual clients. If even a single client contributes inaccurate or deliberately misleading data, it can affect the overall model's integrity and lead to false predictions and compromised performance. In the meantime, biased result is also misinformation. With biased data source, the federated learning could mislead the decision-making with disparate outcome. Detecting and mitigating misinformation in federated learning is challenging as they require effective mechanisms to validate the data and ensure the trustworthiness of the clients' contributions. Strategies like data filtering, client reputation scoring, and robust aggregation methods are employed to address this issue and safeguard the accuracy and reliability of the federated model. Ensuring the integrity of the data in federated learning is crucial to prevent the propagation of misinformation and to maintain the model's credibility and effectiveness in real-world applications.

## *5.3   AI Ethics*

AI ethics play a crucial role in the context of federated learning, where data from multiple sources is aggregated to train a global model. As federated learning involves sensitive data from diverse clients, ethical considerations are paramount to safeguard privacy, security, fairness, and transparency. Even though the well-trained federated learning models can perform decision by strictly following the statistical distribution of the training data, there is no guarantee on the corresponding negative influence to the society. For example, due to high hospital costs, poor people may refrain from seeking medical attention for certain serious illnesses, which could lead AI to believe that such diseases do not exist in certain populations. This is because relevant training data may also be absent [18]. Therefore, AI ethics involves accountability for the actions of the global model and understanding its potential impact on society. By adhering to ethical guidelines and promoting responsible AI practices, federated learning should harness the power of collective intelligence while upholding moral principles and social values.

## 5.4 Responsible and Equitable AI

Responsible and Equitable AI represent another important property in the context of federated learning. Responsible AI can be achieved by ensuring privacy, security, and trust in the context of federated learning. We have discussed privacy and security issues in federated learning, and trust is another important and yet complex security property. Trustworthiness in federated learning involved ethics, ability to mitigate misinformation, biases, and the negative impact of data skewness. Furthermore, equitable AI is another important trustworthiness property in federated learning. It refers to the fairness of federated learning with respect to heterogeneous clients, including those clients with insufficient computing resources to run full-size AI models. One solution approach to ensuring equitable AI in federated learning is to support federated learning with heterogeneous clients, allowing vertical and horizontal partitioning of a global model, to enable clients with insufficient computing resources to participate in (and benefit from) federated learning [33, 91].

## 6 Conclusion

In this chapter, we revealed the truths and pitfalls of understanding two dominating threats: training data privacy intrusion and training data poisoning attack. We formulated the training data leakage attacks based on the intrinsic relationship between the training examples and their gradients. We characterized the training data poisoning attacks based on the attack goals and the poisoning mechanism. We gave a brief overview of the representative defense methods proposed to date and analyzed their pros and cons based on our empirical observations. We conjecture that this study will provide a road map for researchers and practitioners engaging in federated learning field to gain an in-depth understanding on privacy and security threats in federated learning and effective privacy protection and security assurance strategies with strong empirical enlightenment.

**Acknowledgments** This research is partially sponsored by the NSF CISE grants 2302720, 2312758, 2038029, an IBM faculty award, a grant from CISCO Edge AI program, and a GTRI Graduate Student Fellowship.

**Disclaimer** Certain equipment, instruments, software, or materials are identified in this chapter in order to specify the experimental procedure adequately. Such identification is not intended to imply recommendation or endorsement of any product or service by NIST nor is it intended to imply that the materials or equipment identified are necessarily the best available for the purpose.